\pdfoutput=1

\documentclass[11pt]{article}

\usepackage[final]{acl}

\usepackage{times}
\usepackage{latexsym}

\usepackage[T1]{fontenc}

\usepackage[utf8]{inputenc}

\usepackage{microtype}
\usepackage{svg}
\svgsetup{inkscapelatex=false}
\usepackage{inconsolata}
\usepackage{array}
\usepackage{graphicx}
\usepackage{kotex}
\usepackage{threeparttable}
\usepackage[utf8]{inputenc} 
\usepackage[T1]{fontenc}    
\usepackage{hyperref}       
\usepackage{url}            
\usepackage{booktabs}       
\usepackage{amsfonts}       
\usepackage{nicefrac}       
\usepackage{microtype}      
\usepackage{xcolor}         
\usepackage{mdframed}
\usepackage{threeparttable}
\usepackage{graphics}
\usepackage{multirow}
\usepackage{rotating}
\usepackage{fontsize}
\usepackage{tabularx} 
\usepackage{array}
\usepackage{caption}
\usepackage{xcolor}
\usepackage{float}
\usepackage{colortbl}
\usepackage{listings}
\usepackage{enumitem}
\usepackage{amsmath}
\usepackage{cleveref}
\usepackage{hyperref}
\usepackage{xspace} 
\usepackage{subcaption} 
\usepackage{comment}
\usepackage{ulem}
\usepackage{geometry}
\usepackage{adjustbox}
\usepackage{footnote}
\usepackage{tablefootnote}
\usepackage{lipsum} 
\usepackage{tabularx}

\newcommand{\mscell}[1]{\cellcolor{teal!20}#1}

\newcommand\blfootnote[1]{%
  \begingroup
  \renewcommand\thefootnote{}\footnote{#1}%
  \addtocounter{footnote}{-1}%
  \endgroup
}

\definecolor{darkgreen}{rgb}{0.0, 0.7, 0.0}

\title{InfiniPot: Infinite Context Processing on Memory-Constrained LLMs}

\author{Minsoo Kim$^{1,2\dag}$\hspace{1em}Kyuhong Shim$^2$
\hspace{1em}Jungwook Choi$^{1\ddag}$\hspace{1em}Simyung Chang$^{2\ddag}$ \\
{$^1$Hanyang University$^\diamond$} \\
{$^2$Qualcomm AI Research$^{\ast}$, Qualcomm Korea YH} \\
{\texttt {\normalsize\{minsoo2333, choij\}@hanyang.ac.kr}} \\
{\texttt {\normalsize\{minkim, kshim, simychan\}@qti.qualcomm.com}} 
}

\begin{document}
\maketitle

\blfootnote{\hspace{-1.8em}$^\dag$ Work done during an internship at Qualcomm Technologies, Inc.\\
$^\ddag$ Indicates co-corresponding authors. \\
$^\diamond$Hanyang University solely downloaded and evaluated the datasets.\\
$^\ast$Qualcomm AI Research, an initiative of Qualcomm Technologies, Inc.} 

\vspace{-8mm}
\begin{abstract}

Handling long input contexts remains a significant challenge for Large Language Models (LLMs), particularly in resource-constrained environments such as mobile devices. Our work aims to address this limitation by introducing InfiniPot, a novel KV cache control framework designed to enable pre-trained LLMs to manage extensive sequences within fixed memory constraints efficiently, without requiring additional training. InfiniPot leverages Continual Context Distillation (CCD), an iterative process that compresses and retains essential information through novel importance metrics, effectively maintaining critical data even without access to future context. Our comprehensive evaluations indicate that InfiniPot significantly outperforms models trained for long contexts in various NLP tasks, establishing its efficacy and versatility. This work represents a substantial advancement toward making LLMs applicable to a broader range of real-world scenarios.

\end{abstract}

\section{Introduction}\label{sec:intro}

Large Language Models (LLMs) have revolutionized the field of Natural Language Processing (NLP) by achieving unprecedented performance across diverse tasks. However, their capacity to handle long input contexts remains notably limited, primarily due to the predefined maximum context length set during pretraining~\cite{huang2023advancing,wang2024beyond}. Moreover, even within these trained context lengths, LLMs often struggle to effectively process and maintain coherence over extended sequences~\cite{peng2023does, li2023loogle, liu2024lost}. 
The challenge becomes further pronounced in resource-constrained environments such as mobile devices, where memory and computational power are significantly limited. In long context scenario, Key-Value (KV) cache size operates as a hard constraint on the available memory, making LLM inference even more challenging. Efficiently processing long sequences is essential for enabling applications such as document summarization and complex question answering directly on edge devices, yet current LLMs fall short of this capability.

Traditional approaches to addressing long context inputs in LLMs include either increasing the memory capacity to match the long context requirements or using streaming inputs to process information incrementally; however, both methods have substantial drawbacks. Specifically, increasing memory to accommodate long contexts is often impractical, especially in on-device scenarios due to hardware constraints. Moreover, current streaming-based solutions typically leverage only recent context for memory saving, thus failing to fully capitalize on the entire input context~\cite{xiao2024efficient,Beltagy2020Longformer,han2023lm}.

Several attempts have been made to manage long contexts in LLMs, exploring techniques like parameter-efficient fine-tuning (PEFT) and recurrent attention mechanisms~\cite{chen2024longlora,hwang2024transformerfam,dong2024get,zhang2024soaring,mohtashami2023landmark,bulatov2024beyond}. These approaches, however, typically require significant memory increments or additional training steps, leading to inefficiencies and becoming impractical. Other methods have tried modifying positional encoding to prevent out-of-distribution issues but do not address the increased computational memory demands~\cite{chen2023extending,peng2024yarn,jin2024llm}.

Additionally, there have been efforts to utilize Key-Value (KV)-cache compression. Existing KV-cache compression methods primarily focus on compressing prior contexts during the generation stage, rather than effectively managing long input contexts with strict memory constraints~\cite{liu2024scissorhands,ge2024model,li2024snapkv}. In Transformer-based auto-regressive LLMs, the attention mechanism assigns scores to each token based on future context. However, in situations where full context is unavailable, identifying critical input parts becomes challenging, often resulting in information loss and suboptimal performance in memory-constrained settings.

To address these challenges, we propose \textbf{InfiniPot}, a novel KV-cache control framework that allows pre-trained LLMs to handle infinitely long contexts within fixed memory constraints. Our model works akin to a virtual ‘pot’ that processes incoming ingredients. When the pot nears overflow, it distills unnecessary parts and retains only the essentials. Similarly, InfiniPot serves as a \textit{‘memory pot’} that processes incoming token sequence. When the number of KV-cache entries approaches its limit, it compresses the current KV-cache context into a smaller set of entries. This consume-and-compress cycle continues, preventing KV-cache overflow.

To achieve this, we introduce \textbf{Continual Context Distillation (CCD)}, an iterative process that effectively obtains and retains essential information through novel importance metrics. The \textit{Catalyst Prompt (CaP)} consists of carefully designed volatile prompts injected right before the KV-cache reaches its limit, providing strong guidance in generating attention scores and approximating the future importance of tokens within a finite context. Meanwhile, \textit{Novelty under Compression (NuC)} score prioritizes new information that the pre-trained model or the previous context has not encountered by assigning higher importance to such novel content. By combining CaP and NuC, our approach can distinguish representative and novel tokens from less critical ones. This enables the management of long contexts efficiently through multiple CCD cycles.

The proposed methodology promises to empower LLMs to handle extended sequences effectively, making them applicable to a broader range of NLP tasks. Our thorough evaluations demonstrate that InfiniPot enables pre-trained models to achieve performance comparable to, or even surpassing, models explicitly trained to handle long contexts on various long-context NLP tasks, all without requiring additional training.

In summary, our key contributions are as follows:
\begin{itemize}
    \item We propose InfiniPot, the first model-agnostic framework that allows pre-trained LLMs to efficiently handle very long contexts within fixed memory requirements, without additional training.
    \item We introduce Continual Context Distillation (CCD), an iterative process that compresses and retains essential information through novel importance metrics, effectively managing long sequences without future context.
\end{itemize}

\section{Related Work}\label{sec:related}

\subsection{Towards Long Context Window in LLMs}

The increasing recognition of the need for Large Language Models (LLMs) to manage extended information retention has led to diverse approaches. Recently released pre-trained LLMs, such as Mistral-v0.2~\cite{Jiang2023Mistral7} with a context window up to 32K, Claude 3.0~\cite{claude3.0} up to 200K, and Gemini 1.5~\cite{reid2024gemini} up to 1M, reflect this importance.

In line with enhancing models for longer context ranges, LongLoRA~\cite{chen2024longlora} fine-tunes LLMs to handle long contexts through parameter-efficient fine-tuning. TransformerFAM~\cite{hwang2024transformerfam} incorporates a feedback attention mechanism, acting as working memory that periodically reintroduces information to enhance retention. While promising, these methods come with the constraint of requiring fine-tuning.


LongLM~\cite{jin2024llm} addresses the challenge of extended contexts window by sparsely modifying the positional encoding to prevent the Out-of-Distribution (OOD) issues with positional information. This adjustment allows pre-trained models to maintain information over longer periods without additional training. However, LongLM does not tackle the increase in computational memory demands that occur with processing long contexts. Thus, despite these advances, these methods highlight the essential need for optimizing both the capacity for long context retention and the efficiency of computational memory usage in LLMs.

\subsection{KV Cache Compression for Computational Memory Overhead Reduction}

Efficient management of the KV cache is crucial in environments with limited computational memory, such as edge devices. Traditional methods like the Longformer~\cite{Beltagy2020Longformer,hutchins2022block} use a Sliding Window Attention (SWA) mechanism to focus on recent tokens, reducing computational overhead. StreamingLLM~\cite{xiao2024efficient} further refines this by including initial tokens in the KV cache as ‘attention sinks’ to stabilize attention, although it continuously discards intermediate tokens, which limits comprehension of long contexts.

H2O~\cite{zhang2023ho} and TOVA~\cite{oren2024transformers} selectively retain tokens with the highest attention scores but discard tokens within a specific window during generation, constraining long-context utility. Similarly, SirLLM~\cite{yao2024sirllm} selects tokens based on cross-entropy, leading all attention heads to share the same entries, which reduces cache diversity. SnapKV~\cite{li2024snapkv} retains critical information throughout the full context marked by instruction prompts at its end but requires processing the entire context before cache compression, which significantly increases the demand for computational memory, making it less feasible in memory-constrained environments.

In summary, while existing methods improve information retention in LLMs, they often require additional tuning and do not adequately address computational memory impacts. Recent KV-cache compression techniques often overlook long context input scenarios and fail to optimize memory usage during compression. Our proposed Continual Context Distillation (CCD) method fills these gaps by distilling essence of the context into a fixed-size cache, enabling comprehensive caching within limited memory resources. Our method further introduces the Catalytic Prompt (CaP) and Novelty under Compression (NuC) scores to ensure core elements of the context are efficiently preserved.

\begin{figure*}[t]
\centerline{\includegraphics[width=1\textwidth]{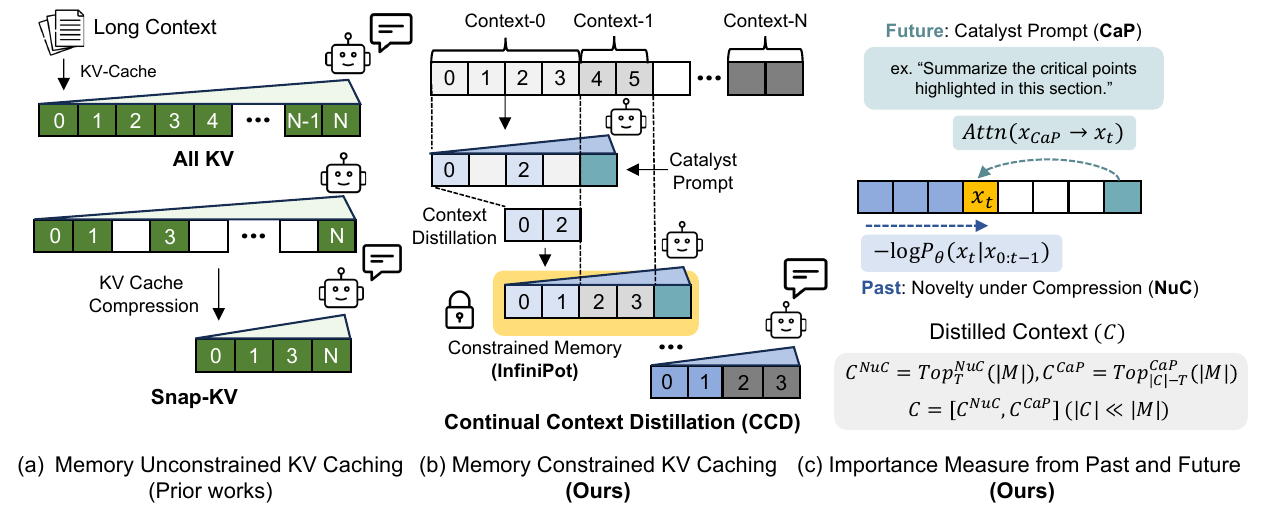}}
\caption{Illustration of KV-cache control methods in long context scenario. (a) Previous memory-unconstrained methods (denoted as SnapKV) showing full context processing. (b) Our memory-constrained KV-cache control using the proposed CCD, where only a limited length of context fits within the memory pot regardless of the total context length. (c) Proposed token importance scoring from perspectives of past and future contexts. Numbers inside the boxes indicate positional indices.}\label{fig:main_illust}
\vspace{-0.1in}
\end{figure*}

\section{InfiniPot: Infinite Context Processing on Memory-Constrained LLMs}\label{sec:method}

In this section, we explain \textbf{InfiniPot}, the proposed KV-cache control framework that enables memory-constrained LLM to handle extremely long contexts.
We first start from the problem definition (Sec.~\ref{ssec:goal}).
Then, we propose a novel context compression method, \textit{Continual Context Distillation (CCD)} (Sec.~\ref{ssec:ccd}).
In particular, the method consists of two unique components, namely \textit{Catalyst Prompt (CaP)} (Sec.~\ref{ssec:catalyst}) and \textit{Novelty under Compression (NuC)} (Sec.~\ref{ssec:novelty}).
Finally, we introduce an effective RoPE position assigning policy that prevents out-of-distribution position problem after KV-cache compression (Sec.~\ref{ssec:rope}).

\subsection{Problem Definition and Objective}\label{ssec:goal}

Our primary objective is to manage long contexts within the strict memory constraints typically found in on-device environments.
It is common to face a hard limit in total amount of memory during on-device LLM inference; for example, DRAM or NPU memory size can become a practical memory cap.
In particular, we focus on the "number of KV-cache entries" kept in memory as a proxy of the actual memory usage.
This can be a good approximation because, assuming the prior KV-cache size used by the model is fixed, the parameter size and activation memory can be considered constant.
Recent studies have also focused on reducing the memory pressure of KV-cache~\cite{shazeer2019fast,ainslie2023gqa,liu2023deja,hooper2024kvquant,liu2024kivi}, as LLM speedup depends on memory bottleneck~\cite{kim2023full,pope2023efficiently,bambhaniya2024demystifying,gholami2024ai}.

We denote the maximum available number of KV-cache entries as $|M|$.
What makes this constraint especially challenging?
First, in this setting, we cannot parallel-process long input over length $|M|$.
Second, we cannot consider long context over length $|M|$ in token generation process.
Please note that this limitation is practically important but surprisingly hasn't been studied much.

\subsection{Continual Context Distillation (CCD)}\label{ssec:ccd}

Before we introduce our methodology, we first review previous KV-cache control methods that overlook memory constraints. In~\autoref{fig:main_illust}(a), all input context KV embeddings participate in generation. For example, SnapKV retains past tokens with high attention scores from context-end prompts, assuming that the entire set of KV embeddings are accessible. During this process, the need to load the entire context input into the model creates a scenario of unconstrained memory usage.

To enforce strictly constrained memory usage, we propose Continual Context Distillation (CCD), a novel method that compresses a long context continuously in a divide-and-conquer manner. Initially, the cache is filled with as many KV embeddings as it can hold (Context-0 in~\autoref{fig:main_illust}(b)). Subsequently, through the Context Distillation process based on our proposed metric, we evaluate the importance of tokens and retain only crucial tokens, reducing the memory size to $|C|$ ($|M|\gg|C|$). $|C|$ is the number of KV-cache entries after CCD. Then, the forward pass computation continues, filling the remaining cache space sequentially (Context-1 in~\autoref{fig:main_illust}(b)). Once the cache becomes full, we repeat the aforementioned process including previously distilled tokens and newly taken ones.
This continual distillation of incoming context manages the KV-cache in a memory-constrained environment, so we call our KV-cache \textit{memory pot}\footnote{One may connect this analogy to the concept of perpetual stew. 
}.

\subsection{Importance Measure from Past and Future}\label{ssec:token}

The key technical innovation of CCD is the importance measure that considers both the past ($x_{0:t-1}$) and future ($x_{t+1:\infty}$) contexts despite a finite cache size with compressed context. We explain CaP and NuC for accomplishing this goal.

\subsubsection{Importance from Future: representative Score from Catalyst Prompt}\label{ssec:catalyst}

At first glance, one can estimate the future importance as "how well the retained context represents the context". Following common practice, this can be formulated using the attention score. For the $t\text{-th}$ token's future importance $u_t$:
\begin{equation}
    u_t = \sum_{i=t+1}^{\infty} \text{Attn}(x_{i} \rightarrow x_{t}),
    \label{eq:importance-future}
\end{equation}
where $\text{Attn}(a \rightarrow b)$ denotes the self-attention probability of how much $a\text{-th}$ token pay attention to $b{\text{-th}}$ token. 
However, Eq.~\eqref{eq:importance-future} sums the contribution of one token to its future tokens, which is not feasible due to the limited future context of memory pot. Therefore, we approximate $u_t$ in finite context by introducing \textbf{Catalyst Prompt (CaP)}. CaP refers to a prompt (i.e., auxiliary context) used solely for importance calculations without being included in the original context.
Via a commodity prompt design, CaP can provide strong guidance to generate an appropriate attention score (see~\autoref{tab:cap_ablation} for the examples of CaP). 
Let the token length of CaP $|P|$, then the approximated future importance $\Tilde{u}_t$ can be represented as below:
\begin{equation}
    \Tilde{u}_t = \sum_{i=|M|-|P|}^{|M|-1} \text{Attn}(x_{i} \rightarrow x_{t})
\end{equation}
See \autoref{fig:main_illust}(c) for a visualization of this computation, where only the attention scores from CaP are used to calculate the actual context. After the attention score is generated, we remove CaP and continue. Note that CaP is always appended at last, right before the pot is about to overflow (see \autoref{fig:main_illust}(b)).
In addition, CaP is calculated per head, so the remaining KV-cache entries may not be synchronized in the token axis after CCD.

\subsubsection{Importance from Past: Novelty Score from Compressed Context}\label{ssec:novelty}

To evaluate the importance of the previous context within memory pot, we propose a new metric called the novelty score from compressed context, namely \textbf{Novelty-under-Compression (NuC)} score. The proposed metric is unique in two aspects. First, the proposed novelty score emphasizes information distinct from the existing context. Second, since CCD already takes into account the representative capability indicated by the attention scores, it would be synergetic to retain differentiating information from the past context. To this end, we quantify the \textit{novelty} $t\text{-th}$ token from the past context using a cross-entropy, $n_t$:
\begin{equation}
    n_t = -\log P_{\theta}(x_t|x_{0:t-1})
    \label{eq:importance-past}
\end{equation}
Although making sense, Eq.~\eqref{eq:importance-past} cannot be directly calculated in CCD setting since not all the past context in memory pot are kept due to continual distillation. To reflect this continuous update, we approximate $n_t$ as follows: 
\begin{equation}
  \Tilde{n}_t=\begin{cases}
    -\log P_{\theta}(x_t|c_{0:|C|-1}; x_{|C|:t-1}), & \text{if $t>|C|$}.\\
    -\log P_{\theta}(c_t|c_{0:t-1}), & \text{otherwise}.
  \end{cases}
\end{equation}
where $c_j$ is the $j\text{-th}$ element in the compressed region.
As opposed to the representative score obtained by CaP, NuC score is per-token based, meaning the two importance scores operate on different axes. Thus, we need additional work to harmonize them.

\subsubsection{Combine Representative and Novelty Scores}\label{sssec:combination}

Representative score ($\Tilde{u}t$) and Novelty score ($\Tilde{n}t$) are complimentary, so we separately keep the tokens with the largest novelty or representativeness among current tokens in the pot ($|C|=|C^{\text{CaP}}| + |C^{\text{NuC}}|$). We prioritize novelty in token selection by allocating $T$ slots within the total cache slots $|C|$ for tokens measured by their novelty scores.
We first select tokens of $\text{Top}_T$ highest NuC scores and fill the slots.
Then, the remaining $|C| - T$ slots are filled with $\text{Top}_{|C|-T}$ tokens, based on the representative scores generated by CaP.
This two-step process ($\Tilde{n}_t \rightarrow \Tilde{u}_t$) prioritizes the coarse-grained novelty score (evaluated per token) before applying the per-head-based representatitveness score. The decision to use two separate metrics is based on their distinct granularity scales and different impacts on token significance. A detailed pseudo code for the token selection process that combines these two scores is provided in the Appendix~\ref{appn:pseudo-code}.

\begin{figure}[t]
    \centering
    
    \begin{subfigure}[b]{\linewidth}
        \centering
        \includegraphics[width=\linewidth]{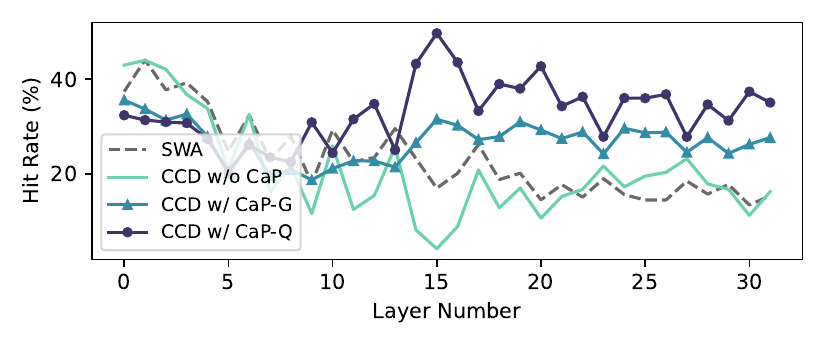}
        \label{fig:first}
        \vspace{-20pt}
    \end{subfigure}
    
    \begin{subfigure}[b]{\linewidth}
        \centering
        \includegraphics[width=\linewidth]{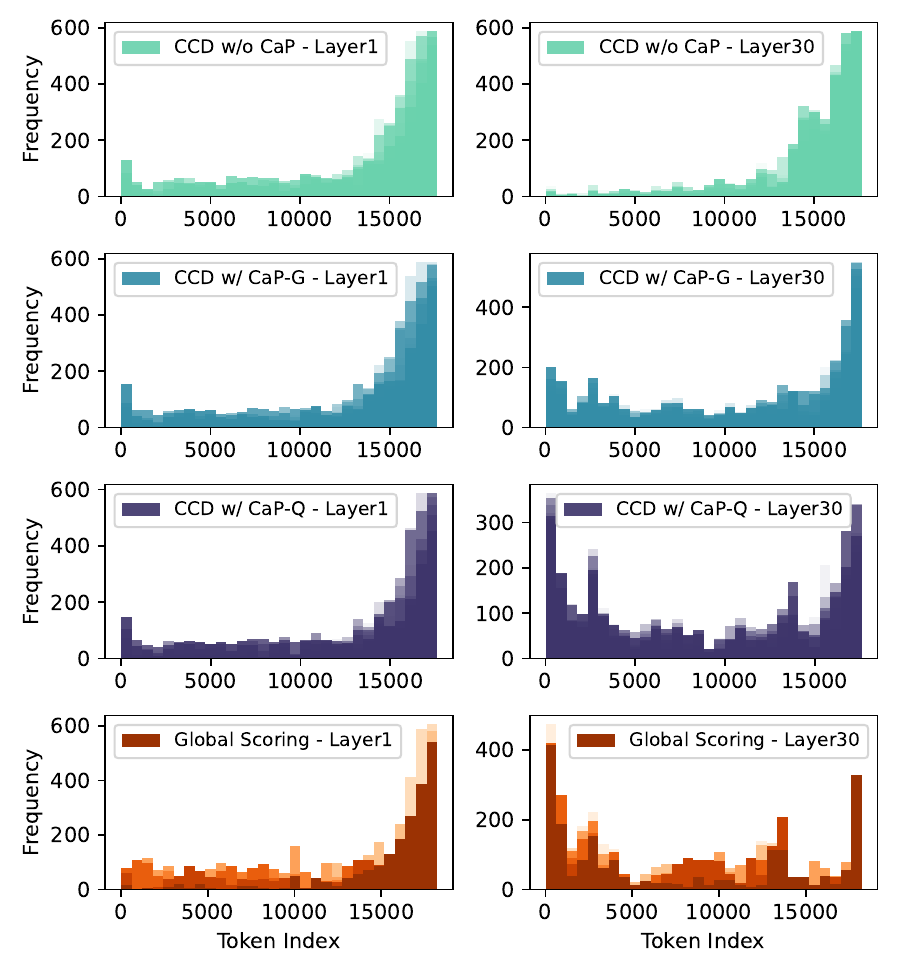}
        \label{fig:second}
        \vspace{-20pt}
    \end{subfigure}
        
    \caption{Top: Hit rate between Continual Context Distillation (CCD) with Catalyst Prompt general (CaP-G) and question (CaP-Q), Bottom: Selected token frequency in memory pot per attention head (left: 1st layer, right: 30th layer) Mistral-inst-v0.3-4K used with HotpotQA task. }\label{fig:cap_analysis}

    \vspace{-0.2in}  
\end{figure}

\begin{figure}[t]
    \centering
    \includegraphics[width=0.95\linewidth]{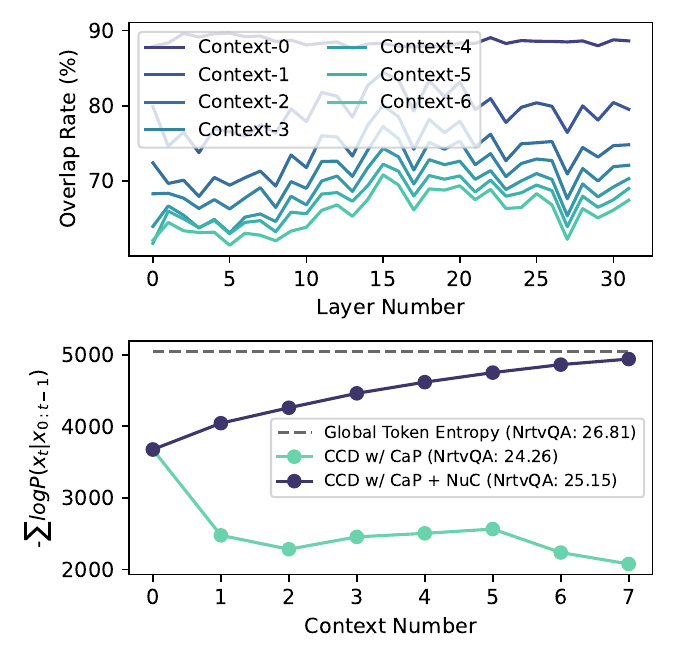}
    \vspace{-0.1in}
    \caption{Top: Comparison of hit rates between CaP and CaP w/ NuP across the CCD cycle. Bottom: Comparison of the summation of selected token's entropy across various CCD configurations (w/ CaP, w/ CaP + NuC). This includes a global token entropy comparison, summing globally selected tokens entropy.}
    \label{fig:nuc_analysis}
    \vspace{-0.1in}
\end{figure}

\subsection{Analysis of Representative and Novelty Scores}\label{ssec:cap_nuc_analysis}
We explore the impact of CaP and NuC scores during the Continual Context Distillation (CCD). To analyze the effectiveness of CaP, we set the memory pot size ($|M|$) to 4K and conduct experiments by either removing CaP at the end of the Context Distillation phase or by altering the prompt type (see \autoref{tab:cap_ablation}). We observe which tokens remained in the memory pot at the end of the context processing. 

\autoref{fig:cap_analysis}(top) presents the Hit Rate analysis, measuring the overlap between tokens attended by the first token generated in an unconstrained memory scenario (where the entire context is accessible for calculating attention scores, called "global scoring") and selected tokens resulting from CCD. Without CaP, CCD's hit rate is similar to Sliding Window Attention (SWA), indicating it struggles to retain crucial tokens. However, CaP significantly improves the Hit Rate, effectively preserving important information.
In \autoref{fig:cap_analysis} (bottom), we show that using CaP results in a closer alignment with global scoring (4th frequency plot), emphasizing the usefulness of CaP in approximating the global representativeness score within a finite context range.
Please see the ablation~\ref{sssec:cap_ablation} for detailed comparative experiments under various prompt designs.

To assess the impact of integrating NuC with CaP in CCD~\footnote{NuC, operating per-token, is not as effective alone because it retains identical tokens across all attention heads.}, we analyze the token overlap rate in the final memory pot when using CaP w/ NuC compared to CaP alone in \autoref{fig:nuc_analysis}(top). As distillation progresses, the overlap between tokens retained using only CaP and those retained with both CaP and NuC decreased, suggesting NuC significantly shifts the token configuration across attention heads.
To further understand this shift, we compare aggregated novelty scores (e.g., token-entropy) of tokens in each attention head's memory pot against global novelty scores, visualized in \autoref{fig:nuc_analysis}(bottom). As more context are processed, novelty scores of tokens retained by CCD increasingly resemble those from global scoring, indicating NuC effectively approximates global perspectives on past importance, even when some past tokens are evicted. This alignment substantially improves performance on NarrativeQA task, approaching the performance of when full past token sets are available. Further qualitative analysis showing which tokens are retained in the memory pot based on CaP and NuC importance scores can be found in Appendix.\ref{appn:qualitative}.

\begin{table*}[t]

\fontsize{18}{25}\selectfont
\setlength{\tabcolsep}{5pt}
\centering
\begin{threeparttable}

\scalebox{0.4}{
\begin{tabular}{llccccccccccccccccc}
\toprule
&\multirow{4}{*}{LLMs-} & \multicolumn{3}{c}{Single-Document QA} & \multicolumn{3}{c}{Multi-Document QA}& \multicolumn{3}{c}{Summarization}& \multicolumn{3}{c}{Few-shot Learning}& \multicolumn{2}{c}{Synthetic} & \multicolumn{2}{c}{Code} & \\

\cmidrule(lr){1-2}\cmidrule(lr){3-5}\cmidrule(lr){6-8}\cmidrule(lr){9-11}\cmidrule(lr){12-14}\cmidrule(lr){15-16}\cmidrule(lr){17-18}
&\rotatebox[origin=c]{}{Method-Memory} & \rotatebox[origin=c]{60}{NrtvQA} & \rotatebox[origin=c]{60}{Qasper} & \rotatebox[origin=c]{60}{MF-en} & \rotatebox[origin=c]{60}{HotpotQA} & \rotatebox[origin=c]{60}{2WikiMQA} & \rotatebox[origin=c]{60}{Musique} & \rotatebox[origin=c]{60}{GovReport} & \rotatebox[origin=c]{60}{QMSum} & \rotatebox[origin=c]{60}{MultiNews} & \rotatebox[origin=c]{60}{TREC} & \rotatebox[origin=c]{60}{TriviaQA} & \rotatebox[origin=c]{60}{SAMSum} & \rotatebox[origin=c]{60}{PCount} & \rotatebox[origin=c]{60}{PRe} & \rotatebox[origin=c]{60}{Lcc} & \rotatebox[origin=c]{60}{RB-P} & Avg. \\
\specialrule{1pt}{2pt}{2pt}

& Avg. Length & 35.29 &	5.79 &	7.90 &	14.98 &	8.36 &	18.15 &	11.67 &	15.87 &	3.05 &	7.59 &	13.80 &	10.88 &	14.17 &	7.58 &	4.25 &	14.68 &	13.19 \\ \midrule

\multirow{8}{*}{\rotatebox[origin=c]{90}{\fontsize{18}{100}\selectfont Memory-Unconstrained}}

&GPT-3.5-PT-16K* & 23.60 & 43.30 & 52.30 & 51.60 & 37.70 & 26.90 & 29.50 & 23.40 & 26.70 & 68.00 & 91.40 & 41.70 & 4.50 & 71.00 & 54.70 & 53.60 & 43.74 \\

& L2-PT-4K & 18.56 & 21.15 &	36.03 &	26.87 &	31.07 &	7.83 &	27.16 &	20.68 &	26.22 &	68.30 &	83.21 &	41.54 &	2.54 & 8.00 & 55.91 & 52.21 & 32.95 \\

& L3-PT-8K & 20.95 & 44.40 & 47.01 & 47.90 & 37.66 & 22.88 & 26.87 & 21.57 & 23.71 & 77.40 & 91.00 & 42.63 & 7.50 & 66.50 & 57.14 & 51.39 & 42.91 \\

& M3-PT-32K & 28.90 & 41.48 &	52.88 &	49.37 &	39.01 &	28.58 &	35.07 &	25.70 &	27.85 &	77.68 &	88.59 &	47.61 &	6.00 & 98.50 &	58.44 &	59.78 &	47.84 \\

\cline{2-19}
&L2-FT-32K*& 16.90 & 27.70 & 41.40 & 31.50 & 20.60 & 9.70 & 30.80 & 22.70 & 26.40 & 63.50 & 82.30 & 34.20 & 1.00 & 30.50 & 53.00 & 55.30 & 34.22 \\

&L3-SE-27K& 27.22 & 32.42 & 43.95 & 49.41 & 40.22 & 27.64 & 28.91 & 24.18 & 26.45 & 77.62 & 90.16 & 41.97 & 3.13 & 89.83 & 57.42 & 50.74 & 44.45 \\


&L3-SnapKV-4K & 23.18 &	31.03 &	41.44 &	44.70 &	37.10 &	23.68 &	26.8 &	23.31 &	26.56 &	77.40 &	91.01 &	42.43 &	6.48 &	67.00 &	57.30 &	51.72 &	41.94\\
&M3-SnapKV-4K & 29.03 &	41.60 &	53.44 &	49.44 &	39.03 &	28.24 &	32.93 &	25.61 &	27.75 &	77.68 &	88.59 &	47.67 &	5.50 &	98.50 &	58.47 &	59.91 &	47.71 \\

\specialrule{1pt}{2pt}{10pt}\specialrule{1pt}{2pt}{2pt}

\multirow{12}{*}{\rotatebox[origin=c]{90}{\selectfont Memory-Constrained}}

& L2-TR-2K & 14.02 &	17.48 &	\textbf{34.30} &	27.01 &	26.91 &	8.14 &	\textbf{27.51} &	20.15 &	\textbf{25.81} & \textbf{64.83} & 76.99 &	\textbf{40.63} & 2.17 &	\textbf{9.00} & \textbf{54.04} & \textbf{50.91} &	31.24 \\

& L2-Streaming-2K & 13.29 &	17.24 &	26.97 &	28.43 &	29.10 &	9.22 &	25.12 &	19.84 &	25.18 &	58.22 &	80.14 &	39.42 &	1.64 & 5.00 & 52.48 & 46.59 & 29.87 \\

& L2-H2O-2K & 13.93 & 15.15 & 26.79 & 31.41 & 28.89 & 8.35 & 24.53 & 20.24 & 25.47 & 61.34 &	\textbf{82.92} & 39.64 & 3.12 &	3.12 &	58.64 &	52.60 &	31.01 \\

& \mscell{L2-InfiniPot-2K} & \mscell{\textbf{17.59}} &	\mscell{\textbf{19.94}} &	\mscell{33.10} &	\mscell{\textbf{32.84}} &	\mscell{\textbf{30.19}} &	\mscell{\textbf{11.13}} &	\mscell{24.78} &	\mscell{\textbf{20.78}} &	\mscell{25.58} &	\mscell{64.52} &	\mscell{74.51} &	\mscell{37.90} &	\mscell{\textbf{3.67}} &	\mscell{3.50} & \mscell{53.42} & \mscell{47.67} &	\mscell{\textbf{31.32}} \\


\cline{2-19}



& L3-TR-4K & 19.74 &	28.61 &	38.34 &	34.29 &	31.94 &	15.98 &	\textbf{27.88} & \textbf{22.91} & \textbf{26.77} & \textbf{75.04} &	88.61 &	\textbf{42.61} &	3.25 &	32.88 &	56.56 &	49.85 &	37.20 \\


& L3-SirLLM-4K & 8.89 & 28.19 & 34.87 & 36.24 & 32.01 & 18.58 & 25.57 & 20.27 & 26.65 & 72.48 & \textbf{91.99} & 40.45 & 5.21 & 18.67 & \textbf{56.86} & \textbf{49.90} & 35.43 \\

& L3-TOVA-4K & 20.89 & 29.10 & 29.47 & 39.76 & 31.49 & 22.61 & 25.35 & 21.89 & 26.18 & 75.12 & 89.52 & 37.41 & 5.66 & 25.43 & 52.38 & 47.89 & 36.26 \\

& \mscell{L3-InfiniPot-4K} & \mscell{\textbf{27.12}} &	\mscell{\textbf{29.26}} &	\mscell{\textbf{43.21}} &	\mscell{\textbf{49.75}} &	\mscell{\textbf{37.21}} &	\mscell{\textbf{26.84}} &	\mscell{26.94} &	\mscell{22.20} &	\mscell{26.72} &	\mscell{73.28} &	\mscell{89.88} &	\mscell{41.34} &	\mscell{\textbf{5.27}} &	\mscell{\textbf{60.25}} &	\mscell{56.42} &	\mscell{47.01} &	\mscell{\textbf{41.50}} \\

\cline{2-19}

& M1-SWA-4K & \textbf{19.47} &	28.47 &	35.63 &	36.23 &	28.58 &	15.24 &	\textbf{27.59} & \textbf{22.61} & 26.16 &	64.11 &	80.02 &	\textbf{19.75} &	2.50 &	\textbf{29.01} &	29.77 &	\textbf{50.43} &	32.22 \\
& \mscell{M1-InfiniPot-4K} & \mscell{18.69} &	\mscell{\textbf{32.86}} &	\mscell{\textbf{42.13}} &	\mscell{\textbf{37.71}} &	\mscell{\textbf{31.25}} &	\mscell{\textbf{17.14}} &	\mscell{26.48} &	\mscell{21.92} &	\mscell{\textbf{26.53}} &	\mscell{\textbf{65.13}} &	\mscell{\textbf{82.26}} &	\mscell{19.65} &	\mscell{\textbf{3.82}} &	\mscell{24.52} &	\mscell{\textbf{52.79}} &	\mscell{46.94} &	\mscell{\textbf{34.36}} \\

\cline{2-19}

& M3-TR-4K & 21.49 &	33.58 &	52.02 &	40.02 &	36.87 &	20.06 &	\textbf{33.13} &	21.51 &	27.82 &	\textbf{75.63} & \textbf{88.76} & \textbf{47.26} & 3.00 &	28.25 &	57.44 &	\textbf{54.31} &	40.07 \\

& \mscell{M3-InfiniPot-4K} & \mscell{\textbf{27.81}} &	\mscell{\textbf{42.75}} &	\mscell{\textbf{53.75}} &	\mscell{\textbf{51.46}} &	\mscell{\textbf{42.94}} &	\mscell{\textbf{28.97}} &	\mscell{32.97} &	\mscell{\textbf{22.46}} &	\mscell{\textbf{27.83}} &	\mscell{72.63} &	\mscell{86.48} &	\mscell{45.36} &	\mscell{\textbf{3.50}} &	\mscell{\textbf{58.89}} &	\mscell{\textbf{57.52}} &	\mscell{52.15} &	\mscell{\textbf{44.22}} \\





\bottomrule

\end{tabular}
}
\vskip -2pt
\caption{LongBench performance comparison for various LLMs and context processing methods. Averaged  length represents the tokenized length using the LLaMA-3 tokenizer. Rows are formatted as LLM - Method - Context Memory, with LLMs denoted as: L - LLaMA, M - Mistral. All LLMs employed are instruction-tuned models (LLaMA-chat/instruct, Mistral-inst) with a size of 7B. Methods include PT (Pre-Trained), FT (Fine-Tuned, LongChat-v1.5), SE (Self-Extend), TR (Truncated) and SWA (Sliding Window Attention). * denotes performance from official LongBench paper, while other results are from our experiments.}

\label{tab:main_longbench}
\vspace{-0.1in}
\end{threeparttable}

\end{table*}

\subsection{Context-Reset Rotary Positional Embedding}\label{ssec:rope}
Handling long contexts requires careful management of positional embeddings to avoid out-of-distribution (OOD) issues~\cite{kazemnejad2024impact}. We introduce Context-Reset Rotary Positional Embedding (\textbf{CR-RoPE}) policy to re-organize positional information across retained tokens with RoPE method. In CCD, key embeddings are stored in the memory pot before applying RoPE in a Pre-RoPE fashion~\cite{xiao2024efficient, hooper2024kvquant}. After each distillation phase, CR-RoPE applies RoPE based on positional indices ranging from 0 to $|C|-1$ to the newly selected entries, effectively preventing OOD compared to memory-unconstrained methods like SnapKV as shown in \autoref{fig:main_illust}(a) and (b). CR-RoPE safeguards any model with a limited context window against OOD issues due to re-organization of positional encoding within the memory-pot size. It also offers efficiency advantages by re-applying RoPE only at the end of each CCD distillation phase, unlike StreamingLLM, which requires re-applying RoPE to the entire set for each single token generated. Specifically, while StreamingLLM may need to recalculate RoPE $L - r - s$ times for a given context length $L$ with $s$ sink and $r$ recent tokens, our method only requires recalculating $L / |M|$ times, making it more efficient in processing long contexts. The performance benefits of CR-RoPE’s reorganization of positional indices will be addressed in Sec.\ref{sssec:nih_with_ccd}, and the latency improvements in Sec.\ref{subsec:efficiency}.

\section{Experiments}\label{sec:exps}

\subsection{Experimental Setup}

We execute our experiments across multiple long context benchmarks under specified memory constraints. In all experiments, the input context length (the number of KV-cache entries) is used as a proxy for actual memory usage.~\footnote{Baseline settings, implementation details, and pre-trained LLMs used in experiments can be found in \autoref{sec:detail}.}

\noindent \textbf{Benchmarks}
We utilize LongBench~\cite{bai2023longbench}, a multi-task benchmark designed for long context understanding, consisting of 6 task categories and 21 diverse tasks covering scenarios such as document QA, summarization, few-shot learning and code completion, with context lengths ranging from 3K to 36K. Additionally, we use the Needle In A Haystack (NIH)~\cite{kamradt2023needle} test to evaluate the ability of models to retrieve critical information from extensive contexts, with varying length from 4K to 1M. 

\noindent \textbf{Baselines} We conduct comparative experiments considering the long context window method and KV-cache compression baselines. For memory-unconstrained scenarios, we considered the recent method, Self-Extend~\cite{jin2024llm} (denoted as SE) and Snap-KV~\cite{li2024snapkv}. For memory-constrained scenarios, we compared the early methood StreamingLLM~\cite{xiao2024efficient} (denoted as Streaming) and H2O~\cite{zhang2023ho}. For recently proposed methods, SirLLM~\cite{yao2024sirllm} and TOVA~\cite{oren2024transformers}, are tested under identical conditions by applying their criteria for retaining critical KV-cache within our proposed CCD pipeline. Furthermore, we utilize the Truncated (denoted as TR) method as a baseline; it is the official LongBench~\cite{bai2023longbench} approach which truncates the middle part of the entire context to fit the predefined context memory.

\subsection{Performance on LongBench}

\noindent \textbf{Memory-Unconstrained} \autoref{tab:main_longbench} demonstrates that the memory-constrained InfiniPot delivers remarkable performance when compared to high-performing, memory-unconstrained LLMs. Specifically, when benchmarking InfiniPot against models such as GPT-3.5-16K and M3-PT-32K, which have high scores of 43.74 and 47.84 respectively, M3-InfiniPot-4K achieves a competitive performance score of 44.22. Additionally, compared with the recent memory-unconstrained technique—Snap-KV, which processes the entire input context before compressing the cache, L3-InfiniPot-4K demonstrates competitive performance, posting a score of 41.50, closely following L3-SnapKV-4K’s score of 41.94.

\noindent \textbf{Memory-Constrained} In conditions where memory is strictly limited, InfiniPot consistently outperforms other methods in handling long contexts. When applying StreamingLLM and H2O, which involve dropping tokens within a restricted window during the long context processing stage, we observe that these approaches yield suboptimal performance (29.87 and 31.01 respectively) compared to the truncation method (TR) which scores 31.24. Furthermore, employing token importance metrics from SirLLM (based solely on token entropy) and TOVA (attention score from the last token) within the proposed CCD pipeline in L3 also resulted in lower performance (35.43 and 36.26 respectively) than TR’s 37.20. In contrast, InfiniPot significantly surpasses all of these baselines across LLaMA and Mistral models. Notably, in M1, where the default SWA\footnote{note that SWA theoretical attention span length is 131K~\cite{Jiang2023Mistral7}} approach is used, InfiniPot also records a higher performance (34.36 vs 32.22). 

Interestingly, as recent LLM models are employed, the efficacy of InfiniPot becomes increasingly highlighted, with performance comparisons in L3 and M3 showing 41.50 and 44.20 against the best baseline scores of 37.20 and 40.07, respectively. Moreover, InfiniPot demonstrates even more remarkable performance improvements with recent LLMs (e.g., LLaMA-3.1/3.2, Gemma-2, and Phi-3), as detailed in the comparative experiments found in Appendix~\ref{appn:recent-llms}

\begin{figure}[t]
    \centering
    \includegraphics[width=\linewidth]{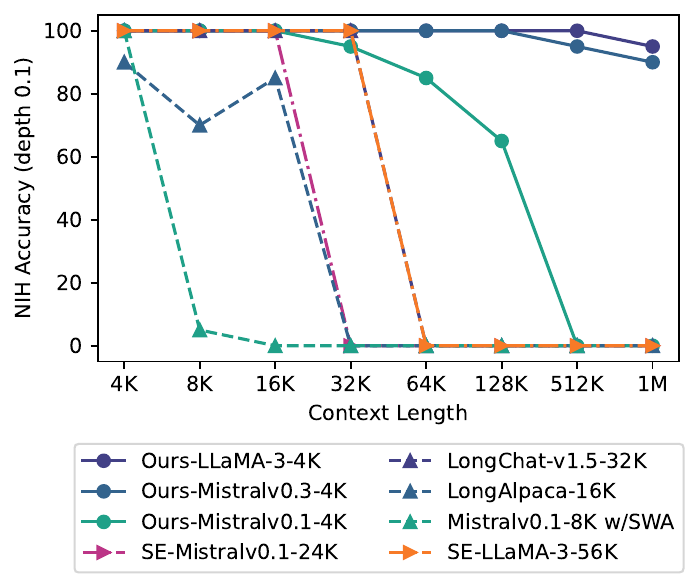}
    \caption{
    Accuracy comparison on the Needle in a haystack (NIH) benchmark at varying context lengths from 4K to 1M. InfiniPot-integrated models (Ours) show superior scalability and maintain high accuracy even at extremely long contexts.
    }
    \label{fig:main_nih}
    \vspace{-0.2in}
\end{figure}

\subsection{Performance on Needle In a Haystack}
\autoref{fig:main_nih} illustrates that our method demonstrates superior scalability and accuracy at extended context lengths in the NIH benchmark. Specifically, InfiniPot with LLaMA-3-4K and Mistral-7B-v0.3-4K (denoted as Ours) maintain high accuracy even at context lengths of 512K tokens. In contrast, models such as LongChat-v1.5-32K and LongAlpaca-16K experience steep performance declines beyond 32K tokens, highlighting their limitations in handling very long contexts.

Our InfiniPot model integrated with Mistral-v0.1-4K also display stable performance up to 128K context length, significantly extending the usable context range while preserving accuracy. Furthermore, models like SE-Mistral-v0.1-24K showed sharp drops in accuracy beyond 32K, indicating that, without InfiniPot, these models struggle with extremely long contexts.

Moreover, our experiments reveal that InfiniPot can handle inputs as long as 1M tokens, demonstrating unparalleled performance not seen in traditionally constrained or pretrained context window extended models. This exceptional scalability underscores the impact of InfiniPot in enabling models to process extraordinarily long context windows efficiently.

\begin{table}[t]

\label{tab:model_performance}


\begin{subtable}{\columnwidth}
\footnotesize
\begin{tabularx}{\columnwidth}{>{\raggedright\arraybackslash}p{2 cm}|X}
\toprule
\textbf{CaP-Type} & \textbf{Prompt Description} \\ \midrule
Placeholder (P) & \texttt{\textbackslash n\textbackslash n\textbackslash n...} \\
Unrelated (U) & \texttt{The sky is blue. The sun is yellow. Here we go. There and back again.} \\
Question (Q) & \texttt{Considering the following question, summarize the critical points highlighted in this section. Question: \{question\}} \\ 
General (G) & \texttt{Summarize the critical points highlighted in this section.} \\ \midrule
General-v1 (G1) & \texttt{Summarize this section.} \\
General-v2 (G2) & \texttt{Highlight the critical points from this section.} \\
\bottomrule
\end{tabularx}
\end{subtable}

\vspace{1em} 
\resizebox{\columnwidth}{!}{    
\begin{tabular}{l|cccc|c}
\toprule
\textbf{LLaMA-3} & {\textbf{QA}} & {\textbf{Summ.}} & {\textbf{FSL}} & {\textbf{Others}} & {\textbf{Avg.}} \\
\midrule
L3-PT-8K       & 35.57 & 24.05 & 70.34 & 45.63 & 42.91 \\
L3-TR-4K       & 28.15 & 25.85 & 68.75 & 35.64 & 37.20 \\
\midrule
L3-CaP-P    & 28.68 & 24.94 & 68.34 & 40.12 & 38.27 \\
L3-CaP-U    & 30.10 & 24.92 & 68.19 & 40.04 & 38.76 \\
L3-CaP-G    & 30.54 & 25.14 & 68.89 & 41.91 & \underline{39.56} \\
L3-CaP-Q    & 35.71 & - & - & - & \textbf{41.50} \\

\midrule
L3-CaP-G1   & 30.49 & 25.01 & 68.52 & 42.20 & 39.48 \\
L3-CaP-G2   & 30.35 & 25.07 & 67.44 & 40.71 & 38.90 \\
\bottomrule
\end{tabular}}
\caption{\textbf{Top}: CaP prompt design description. \textbf{Bottom}: Performance comparison of the LLaMA-3-8B-instruct model with 4K memory on the LongBench tasks.}
\label{tab:cap_ablation}
\vspace{-0.2in}
\end{table}


\subsection{InfiniPot Analysis}\label{ssec:ablation}

In this section, we analyze the individual and combined effects of CaP and NuC on performance with LongBench scores. We will also perform an ablation study to evaluate the impact of each CCD component—CR-RoPE, CaP, and NuC—on performance, using the NIH benchmark. Due to space constraints, we will report average scores for tasks grouped by LongBench categories: Document QA, Summarization, Few-Shot Learning, and Others (Synthetic and Code).

\subsubsection{Catalyst Prompt Design Exploration}\label{sssec:cap_ablation}

We explore various prompt designs in the CaP to approximate the importance score from future contexts, assessing each prompt's impact on LongBench score as detailed in \autoref{tab:cap_ablation}. The proposed CaP-G employs a general instruction to summarize important information, effectively retaining critical parts of the context in a query-agnostic manner. The efficacy of CaP-G is evident, achieving higher LongBench scores compared to CaP-P and CaP-U, which are prompt designs not aligned with CaP’s objective. To further assess the robustness of CaP to prompt design, we employ two variants of the prompt (CaP-G1 and G2), confirming that they achieve similar performances to CaP-G.

In QA tasks where specific queries are predefined, incorporating the question into the CaP prompt leads to additional performance improvements (from 30.54 to 35.71). This enhancement confirms that query-aware context compression via CaP is more effective when queries are present. In general long context scenarios where queries are not pre-accessible, CaP-G still effectively achieves query-agnostic context compression. In \autoref{tab:main_longbench}, we utilize CaP-Q for the QA task and CaP-G for all other tasks.

\begin{table}[t]
\centering

\resizebox{\columnwidth}{!}{    
\begin{tabular}{l|cccc|c}
\toprule
\textbf{NuC Ratio (α)} & \textbf{QA} & \textbf{Summ} & \textbf{FSL} & \textbf{Others} & \textbf{Avg.} \\
\midrule
w/o CaP, NuC     & 28.89 & 24.47 & 67.35 & 32.84 & 36.26 \\ \midrule
0\% (CaP only)      & 33.35 & 25.04 & 68.11 & 39.68 & 39.89 \\ \midrule
25\%                & 33.71 & 25.19 & 68.65 & 41.64 & 40.65 \\
50\%                & 35.71 & 25.14 & 68.89 & 41.91 & \textbf{41.50} \\
75\%                & 35.02 & 25.13 & 68.12 & 41.86 & 41.08 \\ \midrule
100\% (NuC only)    & 26.46 & 24.16 & 68.31 & 32.66 & 35.43 \\
\bottomrule
\end{tabular}
}
\caption{Ablation study of CaP and NuC Components in the CCD Pipeline with LongBench score. LLaMA-3-8B with 4K memory used.}
\vspace{-0.2in}
\label{tab:cap_nuc_ablation}
\end{table}

\subsubsection{NuC and CaP Ablation Study}\label{sssec:nuc_and_cap}
\autoref{tab:cap_nuc_ablation} shows the results from incorporating and removing the CaP and NuC components within the CCD pipeline, as well as comparing performances when these components are combined in various ratios($\alpha$). Including CaP within the CCD significantly enhances performance, as evidenced by an increase from 36.26 to 39.89. In contrast, employing only NuC results in a performance decline from 36.26 to 35.43. This drop is attributed to NuC’s per-token based novelty score which, when used alone, leads to uniformity across all attention head token's entries, reducing per-head diversity and adversely affecting performance.

By harmonizing the CaP and NuC scores as proposed, the combination yields a significant performance gain over using CaP alone, increasing from 39.89 to 41.50. The proportion of NuC tokens ($\alpha$) is treated as a hyper-parameter, and we use a 50\% ratio in our experiments. Our approach clearly demonstrates the effectiveness of combining these two metrics, further highlighting their utility within the CCD framework through comparative results.

\begin{figure}[t]
    \centering
    \includegraphics[trim= 0 5 0 0, clip, width=1\linewidth]{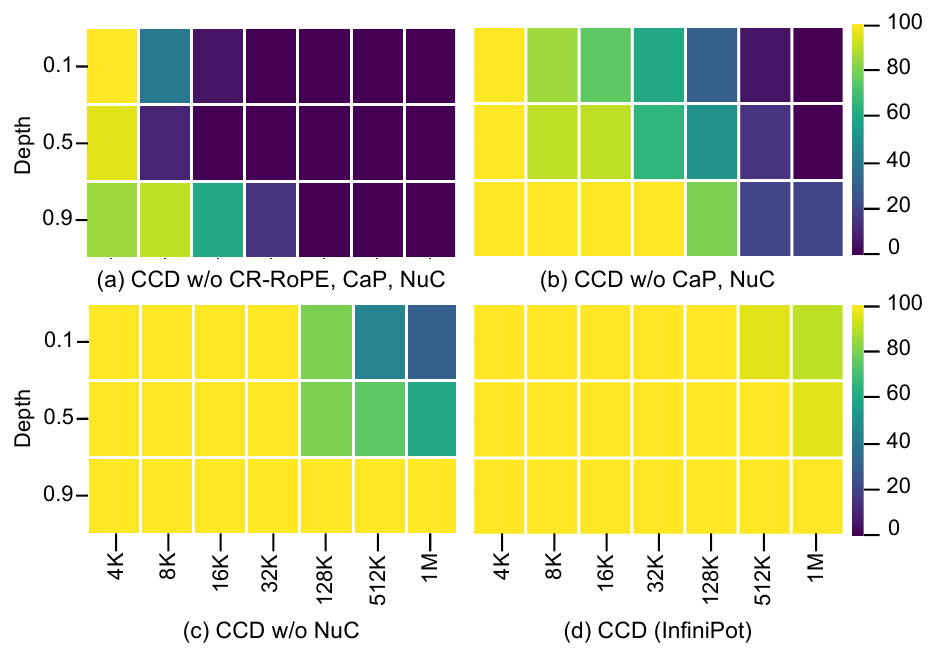}
    \caption{Retrieval accuracy of Mistral-7B-v0.3-4K for the Needle in a Haystack (NIH) passkey task across
varying context lengths from 4K to 1M. The task involved hiding a passkey at different depths (start, middle, end
corresponding to depths 0.1, 0.5, 0.9) and measuring retrieval accuracy as the context length increased. }
    \label{fig:nih_ablation}
\vspace{-0.1in}
\end{figure}

\begin{figure}[t]
    \centering
    \includegraphics[width=1\linewidth]{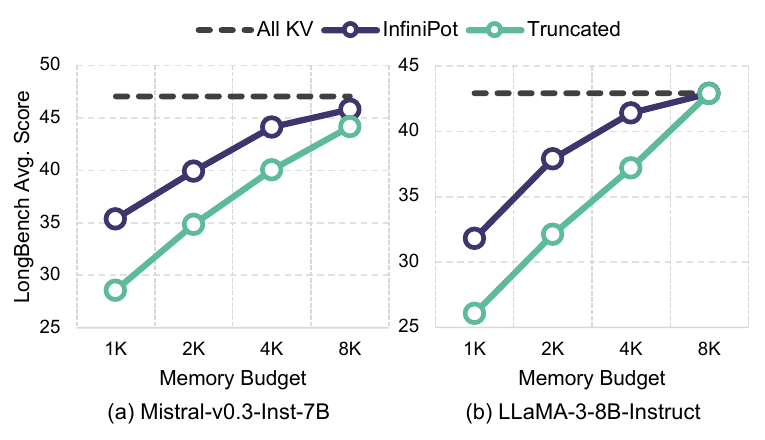}
    \caption{Average LongBench score across memory budget. The maximum memory usage for handling all key-values (All KV) is 32K for Mistral-v0.3 and 8K for LLaMA-3.}
    \label{fig:memory-budget}
\vspace{-0.3in}
\end{figure}

\subsubsection{NIH Ablation Study}\label{sssec:nih_with_ccd}
To explore how each component of CCD enhances performance in the NIH task, we conduct a systematic evaluation of each component’s impact over a range of context lengths from 4K to 1M, as shown in \autoref{fig:nih_ablation}. Initially, \autoref{fig:nih_ablation}(a) presents the results from CCD without incorporating CR-RoPE, CaP, or NuC; this setup merely fills the memory pot sequentially with parts of the context. Here, the performance deteriorates for context lengths shorter than the 32K context window of the employed Mistral-7B-v0.3 model. However, simply reorganizing RoPE within the memory pot significantly recovers performance to match the original model’s context window capabilities, underscoring CR-RoPE’s crucial role in CCD.

Progressing to \autoref{fig:nih_ablation}(c), the addition of CaP to CCD enables handling context lengths up to four times longer than before, enhancing NIH task performance. Further incorporation of NuP, as shown in \autoref{fig:nih_ablation}(d), allows the memory-constrained model to maintain high NIH accuracy even at 1M context length. These results demonstrate how each component of CCD synergistically works to extend the conventional 32K context window by over 30 times, achieving significant context length extension using only 4K of limited memory. Detailed NIH score for each ablation study can be found in \autoref{tab:nih_ablation}.

\begin{figure}[t]
    \centering
    \includegraphics[trim= 0 5 0 0, clip,width=1.1\linewidth]{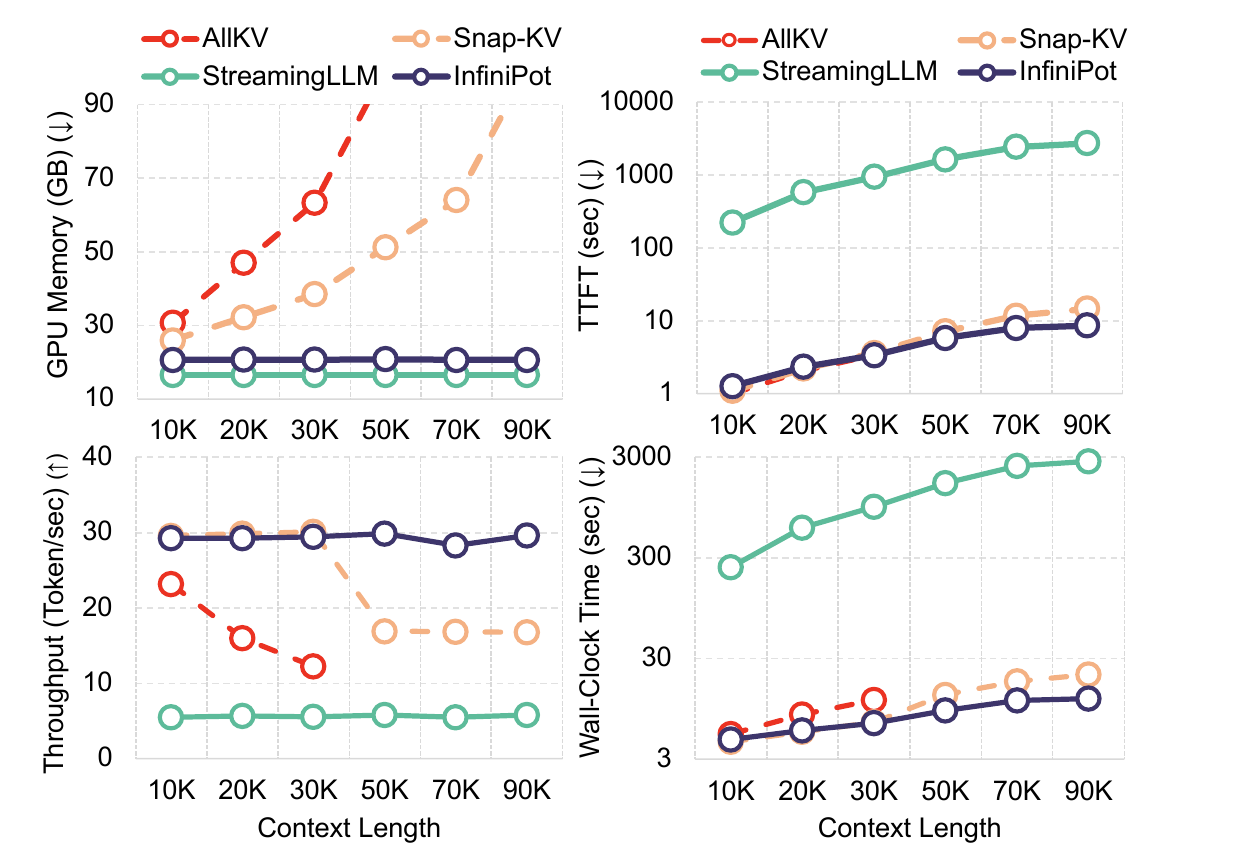}
    \caption{Comparison of memory usage and latency measurements in prefill and generation stages for long context-based generation (GovReport with Mistral-7B-v0.3) under memory-unconstrained (AllKV, Snap-KV-4K) and constrained (StreamingLLM, InfiniPot-4K) settings.}
    \label{fig:memory_latency}
\vspace{-0.2in}
\end{figure}

\subsubsection{Performance Across Memory Budget}\label{sssec:memory_budget}
In \autoref{fig:memory-budget}, we compare the performance of the Truncated (TR) method and our InfiniPot across increasing memory budgets from 1K to 8K on LongBench scores. Both methods exhibit an upward trend in performance as memory budget sizes increase, yet InfiniPot consistently outperforms the baseline across all memory size. This demonstrates InfiniPot’s superior ability to efficiently distill and retain essential context information within the given memory constraints, effectively utilizing the memory budget to achieve enhanced performance in long-context scenarios, even under strict memory constraints.

\subsection{Efficiency Analysis: Memory and Speed Measurement}
\label{subsec:efficiency}

To assess the efficiency of InfiniPot, we evaluate latency and memory usage during text generation with long context inputs under both memory-unconstrained and memory-constrained scenarios as shown in \autoref{fig:memory_latency}. We focus on the memory consumption (GB), latency of Time-To-First-Token (TTFT) in the prefill stage (sec), generation throughput (Tokens/sec), and overall wall-clock time (sec) measurement.

AllKV, caching all KV embeddings, shows a sharp rise in memory use and reduced throughput with increasing context length, indicating performance degradation in memory-bound problem. Snap-KV, loading the entire long context for KV-cache compression, also suffers from increased memory demands during the prefill stage and reduced throughput due to inefficient cache access in long context. In contrast, StreamingLLM, with its fixed cache size, exhibits minimal memory use but suffers from high latency due to the repetitive processing of single token in the long context and the recomputation of RoPE during generation.

InfiniPot stands out by offering consistent performance regardless of context length, optimizing memory use and maintaining high throughput. It matches or surpasses Snap-KV’s speed in the prefill stage, particularly in contexts over 50K, and displays the best token throughput and shortest wall-clock time during generation. This highlights InfiniPot’s superior efficiency and capability in handling long contexts within memory-constrained environments. Detailed latency and memory measurements can be found in Appendix.~\ref{appn:memory_and_latency}

\section{Conclusion}\label{sec:con}

In this paper, we addressed the challenge of enabling Large Language Models (LLMs) to handle long input contexts efficiently in memory-constrained environments. We proposed InfiniPot, a novel KV-cache control framework that uses Continual Context Distillation (CCD) to iteratively compress and distill essential information, even without access to future context. Our evaluations demonstrated that InfiniPot-equipped LLMs can manage extended sequences effectively, achieving performance comparable to or surpassing models explicitly trained for long-context tasks. This work significantly extends the capabilities of pre-trained LLMs, making them more versatile and applicable to a broader range of NLP tasks without requiring additional training.

\section{Limitations}\label{sec:limit}
Despite its promising results, InfiniPot has several limitations that merit further investigation. The current implementation of CCD relies on a predefined compression ratio, which may not be optimal for all types of input data. Future work could explore adaptive compression techniques that dynamically adjust based on context importance. Furthermore, although InfiniPot effectively manages context length within memory limits, its ability to preserve very long-term dependencies across compressed contexts has not been exhaustively tested. Future studies should investigate how well the retained information captures essential long-term dependencies in diverse perspectives. 

Additionally, even though our method is designed with on-device constraints, it has not yet been evaluated in actual on-device environments. Future work should include comprehensive testing on various mobile and edge devices to verify its practical applicability and efficiency under real-world conditions.

\section{Ethics Statement}\label{sec:ethics}
Our research on InfiniPot and the Continual Context Distillation (CCD) algorithm aims to enhance the efficiency and applicability of Large Language Models (LLMs) in memory-constrained environments. We acknowledge several ethical considerations and potential societal impacts stemming from our work.

First, the application of LLMs, particularly in mobile and on-device settings, can involve sensitive user data. Ensuring that our framework complies with data privacy regulations and promotes secure data handling practices is paramount. Additionally, LLMs can inadvertently perpetuate or amplify biases present in training data. While our work focuses on memory efficiency, it is essential to continuously evaluate and mitigate biases in language model outputs to promote fairness and safety~\cite{liu2023trustworthy,gallegos2024bias,bianchi2024safetytuned}.

Our research aims to make advanced NLP technologies more accessible to a wider audience, including those with limited computational resources, aligning with the broader goal of democratizing AI and ensuring equitable access to technological advancements. Moreover, reducing memory usage can contribute to lower energy consumption and, consequently, the environmental footprint of deploying LLMs, particularly in large-scale applications. This consideration is integral to promoting sustainable AI practices~\cite{luccioni2023estimating,stojkovic2024towards}.

We remain committed to addressing these ethical concerns in our ongoing work and encourage the community to engage in discussions about the responsible development and deployment of NLP technologies.

\normalem
\bibliography{custom}


\clearpage
\appendix

\section{Experimental Details}\label{sec:detail}

\paragraph{Implementation} We conduct our experiments and reproduce baseline performances using the PyTorch-based HuggingFace code. The implementation of InifiPot follows the SnapKV codebase\footnote{https://github.com/FasterDecoding/SnapKV}, employing Transformers package version 4.44~\footnote{https://github.com/huggingface/transformers} and FlashAttention2 version 2.6.3~\footnote{https://github.com/Dao-AILab/flash-attention}. All experiments are performed on an A100-80GB GPU. Performance evaluations are conducted in alignment with the official LongBench evaluation procedures, as outlined in their Git repository~\footnote{https://github.com/THUDM/LongBench}. NIH experiments also evaluated following their respective official procedures. In generation stage, all experiments follow greedy decoding strategy. In addition, we adjust the prompts to fit the format used by each LLM, including the system prompt, and provide them as input context.

\paragraph{Baseline} “Memory-Unconstrained” scenarios are where models utilize their maximum trained context lengths, while “Memory-Constrained” scenarios involve models operating with a predefined KV-cache size, such as 2K or 4K, to test performance under restricted memory conditions.

\paragraph{Memory-Unconstrained} For memory-unconstrained scenario, we assess LongChat, finetuned for extended contexts, and include pre-trained LLMs - LLaMA-2-7B-chat, LLaMA-3-8B-Instruct, Mistral-7B-inst-v0.2 and v0.3 supporting contexts from 4K to 32K tokens. For the implementation of LongLM (Self-Extend), we employ three sets of hyper-parameters (group size, window size) in $\{(2, 1024), (2, 1536), (4, 2048)\}$ sets to achieve target context lengths of 27K and 16K, respectively, reporting the configuration with the highest performance.

\paragraph{Memory-Constrained} In the memory-constrained scenario, our baseline includes streaming inference methods such as SWA, StreamingLLM, and H2O. StreamingLLM utilizes 4 start tokens per the official Git implementation, while H2O allocates 4 Heavy Hitter tokens and the remainder to recent tokens, following its default settings from the git repository. In these long context scenarios, these methods process a subset of the long context equivalent to the predefined KV-cache size at once, then handle each subsequent token one-by-one, evicting one old cache entry for each new token to maintain the fixed cache size. 

Another baseline in the memory-constrained setting is the Truncated (TR) method. This approach follows the LongBench official evaluation code~\footnote{https://github.com/THUDM/LongBench/blob/main/eval.py}, where the pre-defined context length is split equally between the beginning and the end of the long context, omitting intermediate information. This method measures performance by providing only the start and end portions of the context. 

\section{Pytorch Style Code for InfiniPot}
\label{appn:pseudo-code}

\lstset{
    language=Python,
    stepnumber=1,
    numbersep=5pt,
    basicstyle=\ttfamily\scriptsize,
    keywordstyle=\color{blue},
    commentstyle=\color{darkgreen},
    stringstyle=\color{red},
    breaklines=true,
    breakatwhitespace=false,
    showspaces=false,
    showstringspaces=false,
    showtabs=false,
    frame=single,
    tabsize=4,
    captionpos=b
}

\begin{lstlisting}[caption={InfiniPot implementation in Pytorch-style}, label={code:InfiniPot}]
1 per_token_ce_loss = torch.nn.CrossEntropyLoss(reduction="none")
2 # Subset of attention scores for CaP (computed during forward pass in attention layer)
3 attn_weights = compute_attn(query_states[:, -len(CaP):, :], key_states, mask)
4 self.attn_cache = attn_weights[..., -len(CaP), :-len(CaP)].sum(dim=-2) 

5 def InfiniPot(outputs, cache_size, NuC_size, attn_cache, label):
   
6     logits = outputs.logits
7     past_key_values = outputs.past_key_values

      # Compute per-token ce-loss
8     token_loss = per_token_ce_loss(logits, label)
9     NuC_indices = token_loss.topk(NuC_size).indices
   
10     # Prioritize NuC-selected tokens
11    attn_cache[:, :, NuC_indices] = attn_cache.max()
   
12    # NuC + CaP 
13    NuC_CaP_indices = attn_cache.topk(cache_size, dim=-1).indices 

      # KV-cache compression
14    key_states_compressed = key_states.gather(dim=2, index=NuC_CaP_indices)
15    value_states_compressed = value_states.gather(dim=2, index=NuC_CaP_indices)
      
      # Update compressed cache
16    past_key_values.update(key_states_compressed, value_states_compressed)
17    return past_key_values
\end{lstlisting}

\begin{table*}[htbp]
\centering
\label{tab:main_nih}
\resizebox{2\columnwidth}{!}{
\begin{tabular}{@{}l|ccccccc@{}}
\toprule
Context Length & 4K & 8K & 16K & 32K & 128K & 512K & 1M \\ \midrule
Passkey Depth & 0.1/0.5/0.9 & 0.1/0.5/0.9 & 0.1/0.5/0.9 & 0.1/0.5/0.9 & 0.1/0.5/0.9 & 0.1/0.5/0.9 & 0.1/0.5/0.9 \\ \midrule 

LLaMA-2-4K & 100/100/100 & OOD & OOD & OOD & OOD & OOD & OOD \\
Ours-2K  & 100/100/100 & 95/95/90 & 100/90/100 & 60/60/90 & 30/40/85 & 25/5/40 & 10/10/25 \\ \midrule
LLaMA-3-8K & 100/100/100 & 100/100/100 & OOD & OOD & OOD & OOD & OOD \\
CCD 4K w/o CaP, NuC  & 100/100/100 & 100/100/100 & 80/80/100 & 80/75/100 & 10/45/70 & 0/5/75 & 15/20/50 \\
CCD 4K w/o NuC & 100/100/100 & 100/100/100 & 100/100/100 & 100/100/100 & 100/100/100 & 100/100/100 & 90/95/100 \\ 
Ours-4K & 100/100/100 & 100/100/100 & 100/100/100 & 100/100/100 & 100/100/100 & 100/100/100 & 95/100/100 \\ \midrule \midrule
Mistral-v0.1-8K w/ SWA & 100/100/100 & 5/100//100 & 0/0/100 & 0/0/100 & OOM & OOM & OOM \\
Ours-4K  & 100/100/100 & 100/100/100 & 100/100/100 & 95/95/100 & 65/40/100 & 0/35/65 & 0/10 \\ \midrule
Mistral-v0.3-32K & 100/100/100 & 100/100/100 & 100/100/100 & 100/100/100 & OOD & OOD & OOD \\
CCD 4K w/o Re-RoPE, CaP, NuC & 100/95/85 & 40/10/90 & 5/0/60 & 0/0/15 & OOD & OOD & OOD \\
CCD 4K w/o CaP, NuC & 100/100/100 & 85/90/100 & 75/90/100 & 60/65/100 & 35/50/80 & 5/15/20 & 0/0/20 \\
CCD 4K w/o NuC & 100/100/100 & 100/100/100 & 100/100/100 & 100/100/100 & 80/80/100 & 45/75/100 &  30/60/100  \\
Ours-4K & 100/100/100 & 100/100/100 & 100/100/100 & 100/100/100 & 100/100/100 & 95/100/100 & 90/95/100 \\
\bottomrule

\end{tabular}}
\caption{Retrieval accuracy of LLaMA and Mistral models for the Needle in a Haystack (NIH) passkey task across varying context lengths from 4K to 1M. The task involved hiding a passkey at different depths (start, middle, end corresponding to depths 0.1, 0.5, 0.9) and measuring retrieval accuracy as the context length increased. Our CCD approach demonstrates the capability to handle the passkey task up to a 1M context length in memory-bounded environments with context range of 4K. OOD denotes positional Out-Of-Distribution, and OOM indicates Out-Of-Memory occurrences during inference on a single A100-80GB GPU}\label{tab:nih_ablation}
\vspace{-0.2in}
\end{table*}


\begin{table}[t]
    \centering
    \resizebox{\columnwidth}{!}{        
    \begin{tabular}{c|ccccc}
        \toprule
        Context & TTFT  & CD time & CD ratio & Gen & Memory  \\
        Length (K) & (sec) & (sec) & (\%) & (sec) &(GB)  \\
        \midrule
        10 & 1.27 {\color{gray}\scriptsize±0.03} & 0.05 {\color{gray}\scriptsize±0.01} & 3.64 & 3.41 {\color{gray}\scriptsize±0.08} & 20.60 \\
        20 & 2.34 {\color{gray}\scriptsize±0.04} & 0.06 {\color{gray}\scriptsize±0.01} & 2.62 & 3.42 {\color{gray}\scriptsize±0.06} & 20.64 \\
        30 & 3.38 {\color{gray}\scriptsize±0.05} & 0.07 {\color{gray}\scriptsize±0.01} & 1.97 & 3.40 {\color{gray}\scriptsize±0.02} & 20.68 \\
        40 & 4.83 {\color{gray}\scriptsize±0.78} & 0.08 {\color{gray}\scriptsize±0.01} & 1.75 & 3.45 {\color{gray}\scriptsize±0.12} & 20.72 \\
        50 & 5.79 {\color{gray}\scriptsize±0.27} & 0.10 {\color{gray}\scriptsize±0.01} & 1.81 & 3.35 {\color{gray}\scriptsize±0.03} & 20.76 \\
        60 & 6.53 {\color{gray}\scriptsize±0.04} & 0.12 {\color{gray}\scriptsize±0.01} & 1.85 & 3.39 {\color{gray}\scriptsize±0.12} & 21.32 \\
        70 & 7.94 {\color{gray}\scriptsize±0.33} & 0.12 {\color{gray}\scriptsize±0.01} & 1.46 & 3.53 {\color{gray}\scriptsize±0.27} & 20.58 \\
        80 & 8.63 {\color{gray}\scriptsize±0.03} & 0.12 {\color{gray}\scriptsize±0.01} & 1.43 & 3.37 {\color{gray}\scriptsize±0.09} & 20.63 \\
        90 & 9.77 {\color{gray}\scriptsize±0.17} & 0.13 {\color{gray}\scriptsize±0.01} & 1.30 & 3.31 {\color{gray}\scriptsize±0.09} & 20.71 \\
        \bottomrule
    \end{tabular}}
    \caption{Results of measuring latency and memory for InfiniPot after GPU warmup. Measurements are taken three times, and both mean and std are reported. "Gen" represents the time taken during the Generation stage, TTFT (Time to First Token) indicates the time during the prefill stage, and CD time reflects the duration spent in Context Distillation (CD), with CD ratio representing its proportion of the total TTFT time.}\label{tab:memory_latency_ours}
    \vspace{-0.2in}
\end{table}

\setlength{\tabcolsep}{10pt}
\begin{table}[t]
    \centering
    \footnotesize
    \resizebox{\columnwidth}{!}{    
 \begin{tabular}{c|ccc}
        \toprule
        Context & TTFT  &  Gen & Memory  \\
        Length (K) & (sec) &  (sec) & (GB)  \\
        \midrule
        10 & 1.09 {\color{gray}\scriptsize±0.02} & 3.38 {\color{gray}\scriptsize±0.09} & 25.95 \\
        20 & 2.21 {\color{gray}\scriptsize±0.01} & 3.34 {\color{gray}\scriptsize±0.09} & 32.19 \\
        30 & 3.62 {\color{gray}\scriptsize±0.01} & 3.32 {\color{gray}\scriptsize±0.08} & 38.45 \\
        40 & 5.27 {\color{gray}\scriptsize±0.03} & 6.01 {\color{gray}\scriptsize±0.12} & 44.83 \\
        50 & 7.15 {\color{gray}\scriptsize±0.01} & 5.90 {\color{gray}\scriptsize±0.08} & 51.24 \\
        60 & 9.32 {\color{gray}\scriptsize±0.01} & 5.92 {\color{gray}\scriptsize±0.09} & 57.66 \\
        70 & 11.75 {\color{gray}\scriptsize±0.02} & 5.93 {\color{gray}\scriptsize±0.04} & 64.08 \\
        80 & 14.69 {\color{gray}\scriptsize±0.04} & 5.96 {\color{gray}\scriptsize±0.15} & 70.88 \\
        \bottomrule
    \end{tabular}}
\caption{Results of measuring latency and memory consumption for Snap-KV. Measurements are taken
three times, and both mean and std are reported.}\label{tab:memory_latency_snap}
\vspace{-0.2in}
\end{table}
\setlength{\tabcolsep}{4pt}

\begin{table}[t]
\centering

\resizebox{\columnwidth}{!}{    
\begin{tabular}{l|cccc|c}
\toprule
\textbf{Recent LLMs} & \textbf{QA} & \textbf{Summ} & \textbf{FSL} & \textbf{Others} & \textbf{Avg.} \\
\midrule
L3.1-PT-128K     & 43.83 & 29.30 & 69.79 & 57.02 & 49.27 \\ 
L3.1-TR-4K       & 34.30 & 26.78 & 68.52 & 39.61 & 40.63 \\
L3.1-InfiniPot-4K & 43.93 & 27.01  & 67.78 & 50.89 & \textbf{46.97}  \\ \midrule
L3.2-PT-128K     & 37.03 & 28.31 & 68.82 & 50.61 & 44.75 \\ 
L3.2-TR-4K       & 31.12 & 26.27 & 66.23 & 33.99 & 37.51 \\
L3.2-InfiniPot-4K & 36.04 & 26.39 & 67.49 & 41.43 & \textbf{41.47} \\  \midrule
G2-SWA-4K         & 38.11 & 24.02 & 69.16 & 45.17 & 43.05 \\
G2-InfiniPot-4K   & 45.41 & 24.46 & 68.42 & 53.74 & \textbf{47.88} \\ \midrule
P3-TR-2K          & 26.78 & 25.34 & 58.76 & 31.50 & 33.69 \\
P3-InfiniPot-2K    & 33.85 & 24.23 & 59.78 & 30.13 & \textbf{35.92} \\
\bottomrule
\end{tabular}
}
\caption{LongBench score with recent LLMs. L3.1 - LLaMA-3.1-8b-Instruct, L3.2 - LLaMA-3.2-3b-Instruct, G2 - Gemma-2-9b-it, P3 - Phi3-3.8b-instruct}
\vspace{-0.2in}
\label{tab:recent-llms}
\end{table}

\noindent\textbf{Calculate Attention Score} As shown in Line 3-4, a subset of attention scores (\texttt{attn\_cache}) are dynamically computed during the forward pass in each layer by summing the scores of CaP tokens attending to previous tokens, serving as the basis for subsequent token selection.\footnote{Since FlashAttention2 is employed in our implementation, we cannot access the entire attention map. To minimize computational overhead, we calculate only the portion of the attention map that represents the scores of previous tokens attended by CaP tokens.}

\noindent\textbf{Calculate NuC Score} In Line 8-9, Using the model’s final logits (\texttt{outputs.logits}), we calculate the entropy for each token (\texttt{token\_loss}). We then select the indices for the top NuC slots (\texttt{NuC\_size}), choosing tokens with the highest entropy.

\noindent\textbf{Combine NuC and CaP} To integrate NuC and CaP effectively, tokens selected based on NuC are prioritized by setting their attention scores the maximum value in the attention score tensor, as described in Line 10-11.

\noindent\textbf{Final Token Selection} As shown in Line 13, Top-K function is employed to determine the indices of the tokens to be retained, set by the total memory size $|C|$ (\texttt{cache\_size}), then extract the necessary key and value embeddings from these indices and update \texttt{past\_key\_values} as described in Line 14-17. Note that, as in Line 4, since the subset of attention scores does not include the scores for the CaP tokens, the CaP tokens are automatically excluded from the cache during the token selection process.

\section{Additional Experimetal Results}

\subsection{NIH Ablation Study Results}
\autoref{tab:nih_ablation} shows the accuracy values from the NIH ablation study discussed in Section~\ref{ssec:ablation}. As outlined in Section.~\ref{sssec:nih_with_ccd}, each components of CCD synergistically enhance NIH accuracy up to a 1M context length.

\subsection{Memory and Latency Results}\label{appn:memory_and_latency}

\autoref{tab:memory_latency_ours} presents the measured latency and GPU memory usage for the InfiniPot method when generating content based on long contexts for the LongBench GovReport task. Experiments are conducted using PyTorch and FlashAttention2~\cite{dao2023flashattention2} without any dedicated kernels. As shown in \autoref{fig:memory_latency}, while the Time to First Token (TTFT) slightly increases with longer context lengths, maintaining the KV-cache in constrained memory ensures consistent generation speeds regardless of context length. Additionally, we measured the Context Distillation time (CD time) during the prefill stage, which accounts for selecting crucial tokens from the KV-cache. This CD time is found to be negligible, constituting only 1 to 3\% of the total TTFT.

\autoref{tab:memory_latency_snap} details the performance metrics for SnapKV, which loads the entire long context into the KV-cache, resulting in significantly increased processing times during the prefill stage as context length grows. This increase becomes particularly severe for context lengths exceeding 50K. Moreover, GPU memory consumption surpasses 70GB at an 80K context length, rendering inference on a single A100-80GB GPU impossible for context length around 100K, even with KV-cache compression. These comparisons underscore InfiniPot’s superior efficiency, demonstrating efficient performance without the need for optimized kernels.

\subsection{LLaMA-3.1/3.2, Gemma-2, Phi-3 Results}
\label{appn:recent-llms}

We conduct comparison experiments with LongBench across recently released LLMs — LLaMA-3.1/3.2~\cite{dubey2024llama3herdmodels}, Gemma-2~\cite{gemmateam2024gemma2improvingopen}, and Phi-3~\cite{abdin2024phi-} — to evaluate the efficacy of InfiniPot, as shown in \autoref{tab:recent-llms}. InfiniPot shows impressive memory-constrained performance, achieving over 6\% improvement on LLaMA-3.1 and nearly 4\% on LLaMA-3.2-3B, a lightweight model released for on-device applications. This highlights InfiniPot’s significant impact, particularly on lightweight LLMs designed for on-device (edge, mobile) environments, where memory is limited.

\begin{table}[ht]
\centering
\small
\begin{adjustbox}{max width=\textwidth}
\begin{tabularx}{.45\textwidth}{@{}p{1.5cm}X@{}}
\toprule
\textbf{Question} & Which \textbf{utility holding company} did Alfred A. \textbf{Marcus} \textbf{works as a consultant?} \\
\textbf{Answer} & Xcel Energy Inc. \\
\midrule
\textbf{Solution} & \\ \midrule \midrule
\textbf{Passage 4} & \dots \textbf{Xcel Energy Inc.} is a U.S. regulated electric utility \dots When H. M. Byllesby began building his \textbf{utility holding company} across the Northwestern region \dots \\ 
\textbf{--> Hop1} & \textit{Xcel Energy is  utility holding company} \\ \midrule
\textbf{Passage 8} & \dots \textbf{Marcus} (born 1950) is an American author and the Edson Spencer Professor of Strategy \dots He has \textbf{worked as a consultant} with companies such as 3M, Corning Inc., \textbf{Xcel Energy, Medtronic, General Mills, and IBM} and has also taught as a visiting professor at Technion \dots \\ 
\textbf{--> Hop2} & \textit{Marcus worked as a consultant with 3M, Corning, Xcel Energy, Medtronic, \dots} \\ \midrule
\textbf{--> Answer} & Xcel Energy \\ \midrule
 \toprule

\textbf{SirLLM} & HotpotQA - 35.94\%\\ \midrule
\textbf{Passage 4} & \sout{Xcel Energy Inc. is a U.S. regulated electric utility \dots When H. M. Byllesby began building his utility holding company across the Northwestern region \dots} \\
\textbf{Passage 8} & Marcus (born 1950) is an American author and the Edson Spencer \sout{Professor of Strategy and Technology} Leadership at the Carlson School of Management, \sout{University of Minnesota, and the Technological Leadership Institute.} He has worked as a consultant with companies such as 3M, Corning Inc.,\sout{ Xcel Energy, Medtronic, General Mills}, and IBM, and \sout{has also taught as a visiting professor at Technion}, INCAE \dots \\ \midrule
\textbf{Answer} & None of the above \\ \midrule \midrule
\textbf{InfiniPot} & HotpotQA - 49.75\% \\ \midrule
\textbf{Passage 4} & \dots \textbf{Xcel Energy Inc}. is a U.S. \sout{regulated} electric utility \dots \sout{When H.} M. Byllesby began building his \textbf{utility holding company} \sout{across the Northwestern region} \dots \\
\textbf{Passage 8} & \dots \textbf{Marcus} (born 1950) is an American author and the Edson Spencer \sout{Professor of Strategy and} Technology \sout{Leadership at the Carlson School of Management, University of Minnesota, and the Technological Leadership Institute.} He \sout{has} \textbf{worked as a consultant} with companies such as 3M, \sout{Corning} Inc.,\textbf{ Xcel Energy}, Medtronic, General Mills, and IBM, and has also taught as a visiting \sout{professor at Technion, INCAE, BI} Norwegian Business School \dots \\ \midrule
\textbf{Answer} & Xcel Energy \\
\bottomrule
\end{tabularx}
\end{adjustbox}
\caption{Qualitative Analysis with HotpotQA test dataset. LLaMA-3-8B-instruct used with 4K memory.}
\label{tab:qualitative_analysis}
\end{table}

\section{Qualitative Analysis}
\label{appn:qualitative}
To further illustrate the effectiveness of our proposed method, we provide a qualitative analysis using a sample from the HotpotQA dataset, as presented in Table~\ref{tab:qualitative_analysis}. The question requires multi-hop reasoning across two passages: (1) identifying the companies where Alfred A. Marcus worked as a consultant, and (2) determining which of these companies is a utility holding company.

Specifically, Passage 8 reveals that Marcus has worked as a consultant with several companies, including Xcel Energy. Passage 4 provides information that Xcel Energy is a utility holding company. To answer the question correctly, both pieces of information must be retrieved and connected.

In the case of SirLLM (Baseline), the cache compression process inadvertently removes critical information. As shown in Table~\ref{tab:qualitative_analysis}, the entire Passage 4 is dropped, eliminating the context that Xcel Energy is a utility holding company. Additionally, key details in Passage 8 about Marcus’s consultancy are also discarded. This leads the model to respond with “None of the above,” an incorrect answer.

Conversely, InfiniPot effectively retains the essential tokens in both passages. The crucial phrases such as “Xcel Energy Inc.” and “utility holding company” in Passage 4, and “Marcus,” “worked as a consultant,” and “Xcel Energy” in Passage 8 are preserved. As a result, the model successfully performs the necessary multi-hop reasoning and correctly answers “Xcel Energy.”

This example demonstrates how InfiniPot’s CaP and NuC work in harmony to prioritize and retain pivotal information within a limited memory context. It enables the LLM to effectively retrieve and reason over long contexts, ensuring that critical tokens are available for accurate comprehension and response generation.

\end{document}